\documentclass{llncs}

\usepackage{graphicx}
\usepackage[T1]{fontenc}
\usepackage{float}
\usepackage{amsmath}
\usepackage{amsfonts}
\usepackage{amssymb}
\usepackage{booktabs}
\usepackage{multirow}
\usepackage{subcaption}
\usepackage{placeins}
\usepackage[colorlinks=true,urlcolor=blue,
            citecolor=black,anchorcolor=black,linkcolor=black,
            unicode]{hyperref}

\begin{document}

\title{Automatic Generation of Executable BPMN Models from Medical Guidelines}

\titlerunning{Automatic Generation of Executable BPMN Models from Medical Guidelines}

\author{
  Praveen Kumar Menaka Sekar\inst{1} \and
  Ion Matei\inst{2} \and
  Maksym Zhenirovskyy\inst{2} \and
  Hon Yung Wong\inst{2} \and
  Sayuri Kohmura\inst{3} \and
  Shinji Hotta\inst{3} \and
  Akihiro Inomata\inst{3}
}

\institute{
  Dept.\ of Mechanical Engineering,
  University of Maryland, College Park, MD 20742, USA\\
  \email{praveenm@umd.edu}
  \and
  Fujitsu Research of America,
  Santa Clara, CA 95054, USA\\
  \email{\{imatei, mzhenirovskyy, awong\}@fujitsu.com}
  \and
  Fujitsu Limited,
  Nakahara-ku, Kawasaki, Kanagawa 211-8588, Japan\\
  \email{\{kohmura.sayuri,hotta\_s,akiino\}@fujitsu.com}
}

\maketitle

\begin{center}
\small\textbf{Project page:}
\url{https://praveen1098.github.io/Automated-BPMN-Generation/}
\end{center}

\begin{abstract}
We present an end-to-end pipeline that converts healthcare
policy documents into executable, data-aware Business Process Model and
Notation (BPMN) models using large language models (LLMs) for simulation-based policy evaluation. We address the main challenges of automated policy digitization with four contributions: data-grounded BPMN generation
with syntax auto-correction, executable augmentation, KPI
instrumentation, and entropy-based uncertainty detection. We evaluate the pipeline on diabetic nephropathy prevention guidelines from three Japanese
municipalities, generating 100 models per backend across three LLMs and
executing each against 1{,}000 synthetic patients.  On well-structured
policies, the pipeline achieves a 100\% ground-truth match with perfect
per-patient decision agreement.  Across all conditions, raw per-patient decision agreement
exceeds 92\%, and entropy scores increase monotonically with document
complexity, confirming that the detector reliably separates unambiguous
policies from those requiring targeted human clarification.
\end{abstract}



\section{Introduction}
\label{sec:introduction}

Healthcare organizations depend on written policy documents to coordinate
patient care, allocate resources, and meet regulatory requirements.  These
documents - often authored in natural language and sometimes spanning
multiple languages - encode the operational logic of programs such as
disease screening, health guidance, and chronic-condition management.
Converting policy text into executable process models that can be simulated
and optimized remains a largely manual task: domain specialists read prose,
construct process diagrams, and keep them synchronized as policies evolve.
This gap between textual policy and computational model limits the scale
and speed of policy analysis, particularly for municipalities operating
dozens of programs simultaneously.

Business Process Model and Notation (BPMN)~\cite{OMG.BPMN.2011} provides a
widely adopted standard for representing processes, and recent large
language models (LLMs) extract structured information from natural
language at scale~\cite{brown2020language}.  However, generating a BPMN
diagram from text constitutes only the first step toward simulation-based policy
assessment.  A practitioner must also make the diagram executable in a
workflow engine, connect it to patient data, instrument it for
performance measurement, and accompany it with quality guarantees - none of
which existing text-to-BPMN methods address in an integrated fashion.

This paper presents an end-to-end pipeline that transforms healthcare policy documents into executable, data-aware BPMN models and
runs simulations to evaluate policy outcomes via key performance indicators
(KPIs).  We frame our investigation around four research questions.

\begin{description}
  \item[\textbf{RQ1} (Scale)] Can an LLM-driven pipeline handle diverse
    policy document formats with minimal manual intervention?
  \item[\textbf{RQ2} (Executability)] Can the pipeline produce BPMN models
    that run directly in a workflow engine without manual augmentation?
  \item[\textbf{RQ3} (Metrics)] Can the pipeline automatically attach KPIs to
    model tasks so that simulation-based policy evaluation becomes possible?
  \item[\textbf{RQ4} (Trustworthiness)] To what extent do automatically
    generated models reproduce the decisions of human-designed models?
\end{description}

To answer these questions, we make four contributions.
First, we develop a structured extraction-and-generation procedure that
enforces BPMN~2.0 compliance through eight named structural rules
checked in a validate-and-repair loop.
Second, we introduce an augmentation step that converts descriptive BPMN
into SpiffWorkflow-executable models~\cite{spiffworkflow}, bridging the
gap between the BPMN standard and a concrete workflow engine.
Third, we automate KPI instrumentation through a majority-vote LLM
procedure that associates predefined KPIs with relevant model tasks,
enabling simulation-based evaluation without manual annotation.
Fourth, we propose an entropy-based uncertainty detector that generates
multiple candidate models per document, simulates each, and flags
documents whose KPI distributions exhibit high variability - exposing
latent ambiguity in the source text for human review.

We demonstrate the pipeline on diabetic nephropathy (DN) prevention
guidelines from three Japanese municipalities, using three LLM backends
(Gemini~2.5 Pro, Gemini~2.5 Flash, and GPT-5.1) and generating 100 models per backend per municipality to characterize generation variability.  DN constitutes a significant public health
concern, accounting for roughly 40\% of new dialysis cases in Japan~\cite{10.3389/fendo.2023.1195167}.
The paper proceeds as follows.  Section~\ref{sec:related} reviews
related work.  Section~\ref{sec:system} presents the system architecture,
including the pipeline stages, KPI-based evaluation framework, and
uncertainty detection.  Section~\ref{sec:experiments} reports experimental
results.  Section~\ref{sec:conclusions} discusses conclusions, limitations, and future work.


\section{Related Work}
\label{sec:related}

Converting natural-language process descriptions into BPMN models has
progressed through three generations of techniques - rule-based,
natural language processing-based, and LLM-driven - each advancing model construction but
leaving execution and quantitative evaluation largely unaddressed.

\textbf{Rule-based and early NLP approaches.}
Friedrich et al.~\cite{friedrich2011automated} introduced one of the earliest full
pipelines, parsing English sentences with Stanford NLP tools and mapping
actors, actions, and control-flow markers to BPMN elements via handcrafted
rules.  Because every mapping is hard-coded, the approach breaks on
non-standard phrasing, domain jargon, or non-English text - all of which
characterize the Japanese policies we study.
Ivanchikj et al.~\cite{10.1145/3365438.3410990} (BPMN Sketch Miner) constrain input to a
lightweight tabular notation, trading coverage for usability but providing
no mechanism for data binding or execution.  The survey by
Bellan et al.~\cite{bellan2021process} found that most of the 10 extraction systems they
reviewed evaluate only structural similarity, with almost no work measuring
decision correctness - a gap our KPI framework closes.

\textbf{LLM-driven model generation.}
LLM capabilities and prompt engineering techniques have dramatically widened
the range of processable texts.
Zirnstein~\cite{zirnstein2024extraction} evaluates three GPT-4-based strategies for
generating BPMN diagrams from unstructured text - direct Mermaid.js
generation, intermediate event-log creation with process mining, and
SAP Signavio JSON export - but stops at structural output without
attempting execution.
Interactive frameworks - K\"opke and Safan's BPMN-Chatbot~\cite{koepke2024bpmnchatbot} and H\"orner's BPMNGen~\cite{horner2025bpmngen} - reduce modeling time but require human
oversight and produce no execution-ready output.
Kourani et al.~\cite{kourani2024promoai} (ProMoAI) wrap GPT-4 with the POWL
representation and pm4py for structural checks, but the generated models
cannot bind to a database schema.
Neuberger et al.~\cite{neuberger2023beyond} improve NER-based extraction recall with a
fully data-driven pipeline that eliminates handcrafted rules, yet the work
targets element extraction rather than end-to-end model execution.
Voelter et al.~\cite{voelter2024multimodal} investigate multimodal GPT-4V capabilities for
extracting process models from combined text-and-image PDF documents,
achieving 87\% similarity with one-shot prompting on a dataset,
but do not attempt execution.
L\^e et al.~\cite{le2025parallelism} augment the PET dataset with 32 new AND-gateway
annotations and train BERT/RoBERTa models for NER with CatBoost for
relation extraction, improving parallelism detection (F1 from 0\% to 23\%
for AND gateways); however, the pipeline produces structural diagrams
without execution semantics.
In our prior work~\cite{matei2026syscon}, we showed that a six-stage
LLM pipeline can reconstruct SpiffWorkflow-compliant BPMN at scale
(similarity $>0.75$ across 387 models), but that study addressed only
generic and simple processes without application-specific KPI evaluation.

A consistent limitation persists across these works: researchers evaluate generated models on
structural fidelity or user satisfaction, never on whether the models
produce functionally-equivalent decisions when executed against data.

\textbf{Towards executable process models.}
Monti et al.~\cite{monti2024nl2processops} (NL2ProcessOps) use LLMs with
retrieval-augmented generation to translate natural language into executable
Python scripts for process engines, but the output is imperative code
rather than a portable BPMN model.
Nivon et al.~\cite{nivon2025givup} (GIVUP) generate BPMN from text and verify
functional properties via LTL model checking, achieving 83\% valid process
generation across 200 descriptions; however, the tool targets formal
verification rather than data-driven execution.
Garg et al.~\cite{garg2025sopstruct} (SOPStruct) convert standard operating procedures
into DAG-based structured representations using LLMs with PDDL-based
evaluation for graph soundness, but bind to no external data source and do
not produce BPMN models.
Bowles et al.~\cite{bowlesjuli} add resource annotations to
healthcare BPMN for workflow optimization, though their annotation process
remains manual.  Complementary to generation,
Shimaoka et al.~\cite{shimaoka2026structure} formulate the optimization of health guidance
decision diagrams as an integer program, finding threshold and notification
assignments that maximize KPIs under budget constraints - assuming the
diagram already exists; our pipeline constructs it from text.

\textbf{Positioning of this work.}
Our pipeline closes two gaps.  First, we automatically validate, repair,
and augment LLM-generated BPMN so that SpiffWorkflow~\cite{spiffworkflow}, a Python-based workflow engine, can execute it directly against a patient database,
with gateway conditions and service tasks grounded in actual schema columns.  Second, we
evaluate models by \emph{functional equivalence}: five KPIs map to tasks
via majority-vote LLM matching and derive from execution traces over 1{,}000
synthetic patients, while an entropy detector distinguishes LLM noise from
genuine policy vagueness.  This KPI-centered evaluation aligns with
healthcare practice and enables simulation-based policy comparison that no prior system
supports.


\section{System Architecture}
\label{sec:system}

This section describes the end-to-end pipeline that transforms policy
documents into executable BPMN models, instruments them for KPI
measurement, and runs simulations for policy evaluation.  We first present a
high-level overview of the pipeline stages and then describe each stage in
detail.

\subsection{Pipeline Overview}
\label{sec:overview}

The pipeline accepts as input a set of PDF policy documents (potentially in a
foreign language) together with a patient database schema that defines the
variables available for decision logic.  It produces as output a collection
of executable BPMN models, each instrumented with KPI measurement points,
along with simulation results comparing generated models against
human-designed baselines.

\begin{figure}[H]
    \centering
    \includegraphics[width=1\linewidth]{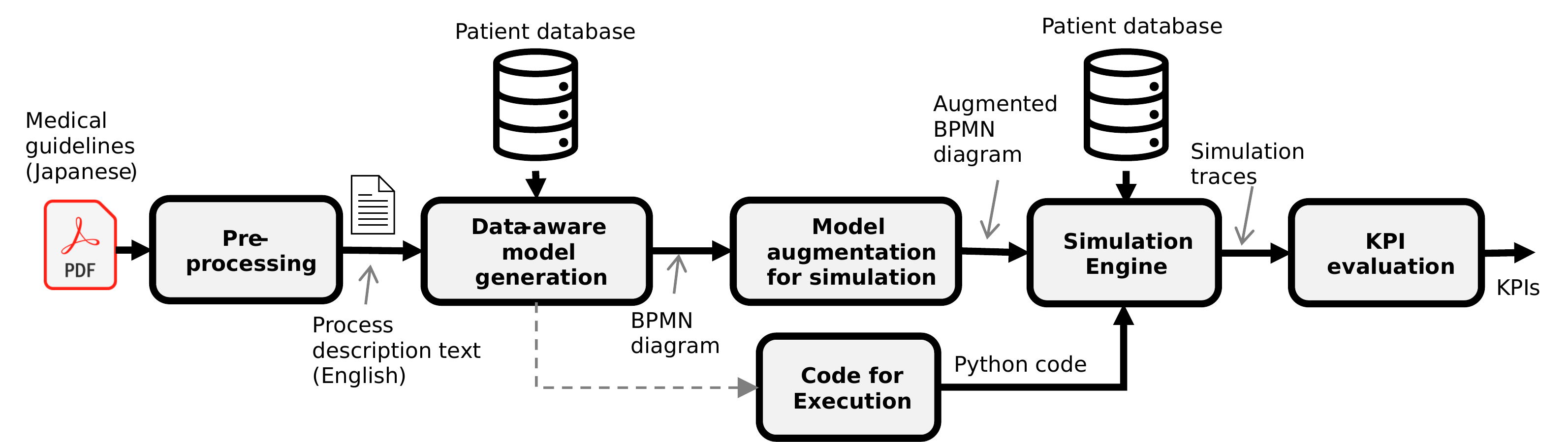}
    \caption{System Architecture for Automated Generation of Executable BPMN Models from Medical Guidelines}
    \label{fig:sys_arch}
\end{figure}

The pipeline comprises six stages executed in sequence as illustrated in Figure~\ref{fig:sys_arch}.

\emph{Stage~1} (Document Processing) extracts text and tables from PDF files
using PyMuPDF4LLM~\cite{pymupdf4llm} and repairs common extraction
artifacts such as broken table layouts and hyphenated word-breaks.
\emph{Stage~2} (Translation) converts non-English content to English while
preserving document structure, numeric thresholds, and logical operators.
\emph{Stage~3} (Narrative Generation) transforms the cleaned English text
into a formal process narrative whose eligibility conditions reference only
columns that exist in the patient database.
\emph{Stage~4} (BPMN Synthesis) converts the narrative into a BPMN~2.0 XML
model and validates it against eight structural compliance rules, repairing
violations automatically.
\emph{Stage~5} (Executable Augmentation) adds engine-specific execution
semantics so that SpiffWorkflow, a Python-based BPMN workflow engine, can run the model.
\emph{Stage~6} (KPI Instrumentation and Simulation) attaches KPI
measurement logic to model tasks, executes the model against synthetic
patient records, and collects performance metrics.

We implement each stage as an LLM-driven module whose behavior we control with a structured
prompt containing explicit validation constraints, complemented by
programmatic validation to minimize LLM inconsistencies.
Generated models use a BPMN~2.0 construct palette of start events, end
events, exclusive gateways (split and merge), service tasks, script tasks,
and conditional sequence flows - sufficient for the decision-centric
eligibility-screening processes common in healthcare and fully supported by
SpiffWorkflow.

\subsection{Document Processing and Translation}

We process source documents with PyMuPDF4LLM to extract text and
tabular content from PDF files, handling multi-column layouts and
hierarchical tables.  A cleanup pass repairs common extraction artifacts
such as vertically rendered text, hyphenated word-breaks, and blank filler
rows.

The cleanup module applies three categories of repair.
\emph{Structural repairs} reconstruct table boundaries that the PDF
extraction breaks across page boundaries, re-merge cells that the extractor
splits vertically, and remove decorative rows (e.g., horizontal rules
rendered as sequences of underscores).
\emph{Typographic repairs} rejoin hyphenated words at line breaks, normalize
whitespace, and replace Unicode look-alikes (e.g., full-width digits) with
their ASCII equivalents to ensure downstream parsability.
\emph{Content-integrity checks} verify that numeric values appearing in
eligibility thresholds (e.g., ``HbA1c $\geq$ 6.5\%'') survive extraction
without truncation or digit transposition; when a value falls outside a
domain-plausible range, the module flags it for manual review rather than
silently propagating an error.

For non-English documents, the LLM translates the cleaned content
to English with explicit instructions to preserve table structure, reference
identifiers, and numeric thresholds.  We supply the LLM with a glossary of
domain-specific terms (e.g., Japanese medical examination codes, municipality
program names) to reduce mistranslation of jargon.  The translation prompt
further instructs the model to retain the original term in parentheses
alongside the English translation whenever the original carries regulatory
significance, enabling downstream auditing against the source document.

\subsection{Data-Grounded Narrative Generation}
\label{sec:narrative}

The central challenge in converting policy text to an executable process
model lies in bridging the gap between natural-language descriptions and formal
decision logic.  A policy document might state, for example, that
``individuals diagnosed with diabetes whose HbA1c is 6.5\% or higher are
eligible for health guidance''.  To convert this sentence into a computable
gateway condition, the system must (a)~identify the relevant clinical
concepts (diabetes diagnosis, HbA1c level), (b)~find corresponding columns
in the patient database (e.g., \texttt{Diabetes}, \texttt{HbA1c}), and
(c)~construct a formal Boolean expression (\texttt{Diabetes == 1 and
HbA1c >= 6.5}).  This mapping must be precise: selecting a wrong column or
inventing one that does not exist in the database produces a model that
silently makes incorrect decisions.

The narrative-generation stage addresses this challenge through a
structured, five-phase prompting process that converts policy prose into a
\emph{database-grounded formal narrative} - a temporally ordered description
of the process in which every condition takes the form of a
Boolean expression over actual database columns.

\emph{Phase~1 (Structural Analysis)} decomposes the policy text into a
sequence of process steps.  The LLM identifies the high-level workflow
stages (e.g., candidate identification, notification, screening, guidance
enrollment) and organizes them into a directed acyclic sequence.  Each step
receives a descriptive label and a preliminary indication of whether it
involves a decision (gateway) or an action (task).  The prompt instructs the
model to preserve the document's own ordering and section structure wherever
possible, reducing the risk of hallucinated control flow.

\emph{Phase~2 (Criterion Tokenization)} extracts the criteria embedded in each decision step and decomposes them into atomic
predicates connected by logical operators.  For example, the clause ``fasting
glucose $\geq126$~mg/dl or HbA1c $\geq6.5\%$'' yields two atomic
predicates (\texttt{fasting\_glucose >= 126}, \texttt{HbA1c >= 6.5})
linked by~\texttt{OR}.  The tokenizer also identifies negation (``excluding patients
with\ldots''), range predicates (``eGFR between 30 and 60''), and set
membership (``urinary protein $\geq\pm$'').

\emph{Phase~3 (Expression-Tree Construction)} applies standard operator
precedence to assemble the token stream into a
hierarchical expression tree.  Parenthesization instructions resolve scope
ambiguities that arise when the source text uses implicit conjunction
(e.g., ``patients with diabetes \emph{and} impaired renal function
\emph{or} proteinuria'' admits two parses).  The prompt supplies the LLM
with the explicit precedence rules and instructs it to default to the
narrower (conjunctive) parse unless the document's formatting (e.g.,
bullet-list nesting) indicates otherwise.

\emph{Phase~4 (Database-Schema Mapping)} maps each leaf node in the
expression tree - representing a single criterion - to one or
more columns in the patient database.  The pipeline ranks candidate columns
by semantic specificity using a three-tier hierarchy:
\begin{enumerate}
  \item \emph{Exact match with qualifying terms.} A column whose name
    encodes both the concept and a qualifier (e.g.,
    \texttt{Type\_2\_Diabetes\_Prior} for ``previously diagnosed with Type~2
    diabetes'') receives the highest rank.
  \item \emph{Exact concept match without qualifiers.} A column that names
    the concept alone (e.g., \texttt{Diabetes}) ranks second.
  \item \emph{Keyword-level match.} A column whose name shares a keyword
    with the criterion (e.g., \texttt{DM\_Treatment} matching ``diabetes
    medication'') ranks third.
\end{enumerate}
When multiple columns share the same rank, the pipeline combines
them with \texttt{OR}. If a policy criterion cannot match any database
column with sufficient confidence, the pipeline omits the corresponding
logic rather than inventing a fictitious variable, ensuring that no
spurious columns enter the model.

\emph{Phase~5 (Expression Generation)} traverses the mapped expression tree
depth-first to produce the final Boolean expression with proper operator
syntax and parenthesization.  The output expression uses Python syntax
(e.g., \texttt{and}, \texttt{or}, \texttt{==}, \texttt{>=}) so that
downstream stages can embed it directly in BPMN gateway conditions without
an additional translation step.

The output of this stage is a formal narrative that reads like a process
description but whose every conditional statement constitutes a database-executable
expression.  This narrative serves as the specification from which the next
stage generates BPMN XML.

\subsection{BPMN Synthesis and Structural Validation}

The synthesis stage converts the formal narrative into a BPMN~2.0 XML
model.  The LLM receives the narrative together with the list of available
database variables and generates the XML directly, including element identifiers,
sequence flows, gateway conditions, and layout coordinates.

We validate every generated model against eight structural compliance rules:
\begin{itemize}
  \item[\textbf{R1}] The model contains exactly one start event with no incoming flows.
  \item[\textbf{R2}] The model contains at least one end event.
  \item[\textbf{R3}] Every task possesses exactly one incoming and one outgoing flow.
  \item[\textbf{R4}] Every gateway functions as either a split (one incoming, multiple outgoing) or a merge (multiple incoming, one outgoing), but not both.
  \item[\textbf{R5}] Every split gateway designates exactly one default outgoing flow.
  \item[\textbf{R6}] All non-default outgoing flows from a gateway carry a Python condition expression with proper XML entity encoding (e.g., \texttt{\&gt;} for~$>$).
  \item[\textbf{R7}] No two outgoing flows from the same gateway target the same element.
  \item[\textbf{R8}] All variables referenced in condition expressions correspond to columns in the patient database schema.
\end{itemize}

When the validator detects a rule violation, the system applies targeted
repairs automatically.  For~R3 violations, the repair module inserts a
merge gateway upstream of elements that receive multiple incoming flows.
For~R4 violations (a gateway acting as both split and merge), the module
decomposes the node into a merge gateway followed by a split gateway
connected by a single sequence flow.  For~R5, the module identifies the
outgoing flow whose condition is most general (fewest conjuncts) and
designates it as the default.  For~R6, the module applies XML entity
encoding to all condition strings.  For~R7, the module removes duplicate
flows and re-routes the second target to a newly inserted intermediate task
that preserves the intended semantics.  For~R8, the module queries the LLM
with the invalid variable name and the database schema to obtain the
closest valid column; if no match exceeds a confidence threshold, the
repair module removes the offending condition entirely and logs a warning.

Validation and repair iterate until all eight rules pass or a maximum
iteration count (default: five) is reached.  Models that still violate any
rule after the final iteration are marked as generation failures and
excluded from simulation.  The synthesis module also performs
activity-level reasoning to ensure that the model contains only executable
tasks: any task whose description references an action outside the BPMN
construct palette (e.g., ``send email'' without a corresponding service
binding) is flagged for review.

\subsection{Executable Augmentation}

The BPMN standard does not prescribe how workflow engines should execute models;
different workflow engines require different runtime annotations.  The
augmentation stage converts the validated BPMN into a SpiffWorkflow-executable
format through three transformation steps.

\emph{Task-type conversion.}  Generic BPMN user tasks, which assume human
interaction, become script tasks or service tasks that SpiffWorkflow can
execute programmatically.  Each converted task receives a Python code
block that reads patient-record fields from the workflow's data context
and writes decision outcomes (e.g., \texttt{eligible = True}) back to
the context for downstream gateways.

\emph{Gateway-condition translation.}  Condition expressions originally
written for BPMN XML compliance (with XML entity encoding and namespace
prefixes) translate into raw Python Boolean expressions that
SpiffWorkflow's expression evaluator can parse.  The augmentation module
verifies syntactic validity by parsing each expression with Python's
\texttt{ast.parse} before embedding it; any expression that fails to parse
triggers a re-query to the LLM with the parse error as feedback.

\emph{Data-context binding.}  The augmented model binds to a pandas-backed
data context that loads patient records at runtime.  Each patient record
populates the workflow's data dictionary, making database columns directly
accessible as Python variables in gateway conditions and script tasks.
The augmentation module verifies that every variable referenced in the
model appears in the data-context schema, providing a second layer of
defense (complementing~R8 from the synthesis stage) against spurious
variable references.

A gateway-preservation invariant governs the entire augmentation process:
the module asserts that the set of gateway conditions (modulo syntactic
normalization) before and after augmentation is identical, ensuring that
the transformation does not alter any decision logic.

\subsection{KPI-Based Evaluation Framework}
\label{sec:kpi}

The final pipeline stage instruments the executable model for quantitative
evaluation and runs simulations.  The goal is to determine whether a
generated model makes the \emph{same decisions} as a human-designed
baseline when both process the same patient data.  We operationalize this
comparison through five domain-specific KPIs derived from the
\textit{Diabetic Nephropathy Aggravation Prevention Program}~\cite{ikeda23}.

These KPIs serve as \emph{functional-equivalence metrics}: if a generated
model produces the same KPI values as the human baseline on the same test
population, the two models are functionally equivalent with respect to the
policy decisions they encode.  The KPIs do not predict
real-world clinical outcomes - that would require validation against actual
patient data, which lies beyond the scope of this work.

The five KPIs capture different aspects of the health guidance workflow:
\emph{Notification Count}~(NC) and \emph{Health Guidance Count}~(HC)
directly count patients receiving notifications and patients who accept guidance sessions,
respectively; \emph{Guidance Resource Utilization}~(RU),
\emph{Health Improvement Rate}~(HI), and \emph{Medical Cost
Savings}~(CS) derive from HC through domain-specific empirical scaling
factors (capacity ratio, improvement-rate coefficient, and
dialysis-prevention cost, respectively) and therefore map to the same
model task.  Because RU, HI, and CS are deterministic functions of
HC, only two independent metrics - NC and HC - underlie the KPI-based
evaluation; we complement this aggregate view with a per-patient
decision-agreement analysis in Section~\ref{sec:per_patient}.

\textbf{KPI-to-task mapping.}
For each KPI, an LLM prompt identifies the model task whose execution
semantics most closely correspond to the KPI definition.  Because LLM
outputs exhibit stochastic variation, we repeat the mapping $k$~times and select the task assignment that appears most frequently
across repetitions via majority vote.  This voting mechanism filters
transient mapping errors while preserving consensus assignments.  The
selected mapping determines where the simulation engine increments each
KPI counter during model execution.

\textbf{Simulation and aggregation.}
We execute each instrumented model against 1{,}000 synthetic patient records
whose clinical attributes match published DN screening statistics.
The synthetic population spans the full range of values for each database
column, including borderline cases near every decision threshold, to ensure
that KPI differences between models reflect genuine logic discrepancies
rather than artifacts of a population that avoids boundary values.
We use the same dataset for both human-designed and generated models, so
any patient-level or aggregate KPI difference reflects a difference in model logic.

\textbf{Uncertainty detection.}
For each policy document, the pipeline generates $M$~narratives followed by candidate BPMN model for each narrative, per LLM backend.  The pipeline simulates every model and computes the
normalized Shannon entropy over the empirical distribution of KPI
combinations:
\begin{equation}
  H_{\text{norm}} = -\frac{1}{\log K}\sum_{i=1}^{K} p_i \log p_i\,,
  \label{eq:entropy}
\end{equation}
where $K$ denotes the number of distinct KPI combinations observed and
$p_i$ denotes the fraction of models producing combination~$i$.
A normalized entropy of 0\% indicates that all models agree on a single
KPI outcome (no ambiguity); a value approaching 100\% indicates near-uniform
disagreement across $K$ outcomes.

The entropy score serves as an \emph{ambiguity detector}: high entropy flags
documents whose natural-language descriptions admit multiple valid
interpretations, directing human reviewers to the specific policies that
require clarification.  The detector also distinguishes between
\emph{LLM noise} (random generation failures that produce all-zero KPIs)
and \emph{genuine policy vagueness} (multiple non-trivial KPI clusters
corresponding to distinct but defensible readings of the source text)
through the shape of the KPI distribution: noise manifests as a single
outlier cluster alongside a dominant mode, whereas genuine vagueness
produces multiple comparably-sized clusters.


\section{Experimental Results}
\label{sec:experiments}

This section evaluates the pipeline on real-world healthcare policies and
reports results addressing the research questions posed in
Section~\ref{sec:introduction}.

\subsection{Experimental Setup}

We evaluate the pipeline on DN health guidance policies from three cities
in Japan, referred to as {\tt City~1}, {\tt City~2}, and {\tt City~3},
spanning a range of structural complexity. For each municipality, we compare human-developed BPMN models (ground
truth) against automatically generated models via KPI-based functional
equivalence rather than structural isomorphism.
We employ three LLM backends spanning two model families: Gemini~2.5 Pro
and Gemini~2.5 Flash (Google, both in \emph{thinking mode}), and GPT-5.1
(OpenAI).  For each backend we generate 100~models per municipality, execute each against 1{,}000
synthetic patient records, and evaluate the five KPIs from
Section~\ref{sec:kpi}.  The pipeline processes all three policies without
manual intervention beyond translation verification.

The 1{,}000 synthetic patient records span the full range of clinical
attributes, including borderline values at each decision threshold
(e.g., HbA1c near 6.5\%, eGFR near 30 and 60), ensuring that KPI
agreement reflects decision-level fidelity rather than an artifact of a
population that avoids boundary cases.

\subsection{Test Cases}
\label{sec:testcases}

The {\tt City~1} policy is well-structured, with a conjunction of two
hierarchically organized criterion groups.  Group~A (diabetes) requires
one of: claims-data diagnosis, current treatment (oral or insulin), or
fasting glucose $\geq126$~mg/dl / HbA1c $\geq6.5\%$.  Group~B
(impaired renal function) requires one of: urinary protein $\geq1+$,
eGFR 30-60, urinary protein $\geq\pm$ with eGFR 60-90, or diabetic
nephropathy history.  Eligibility requires A~$\wedge$~B.  Exclusion
criteria remove patients with type~1 diabetes or cancer.

The {\tt City~2} policy targets Stage~2 or Stage~3 DN patients with
a compound Boolean eligibility structure
($(\text{\textcircled{1}} \vee \text{\textcircled{2}}) \wedge
 (\text{\textcircled{3}} \vee \text{\textcircled{4}})$), where
$\text{\textcircled{1}}$/$\text{\textcircled{2}}$ cover diabetes criteria (fasting
glucose or HbA1c thresholds, or prior diagnosis with treatment) and
$\text{\textcircled{3}}$/$\text{\textcircled{4}}$ cover renal criteria (urinary
protein/albumin levels or eGFR 30-60).  The same exclusion criteria
apply.  Guidance spans a structured six-month program with pre- and
post-intervention comparison.

The {\tt City~3} policy is the most complex, featuring implicit
sequencing and temporal dependencies across two fiscal years.
Stage~$\text{\textcircled{1}}$ determines eligibility for an albumin test
based on prior-year checkup results (HbA1c 6.0-6.4\%, urinary
protein $(-)$ to $(+)$, no diabetes medication, ages 60-64).
Stage~$\text{\textcircled{2}}$ determines eligibility for a guidance
interview based on Stage~$\text{\textcircled{1}}$ results or separate
clinical criteria (urinary protein $(\pm)+$, eGFR 30-60, HbA1c
$\geq6.5\%$).  The temporal coupling is implicit, blurring the
boundary between selection criteria and fallback procedures.

\begin{figure}[htbp]
  \centering
  \begin{subfigure}[t]{\textwidth}
    \centering
    \includegraphics[width=\textwidth]{%
      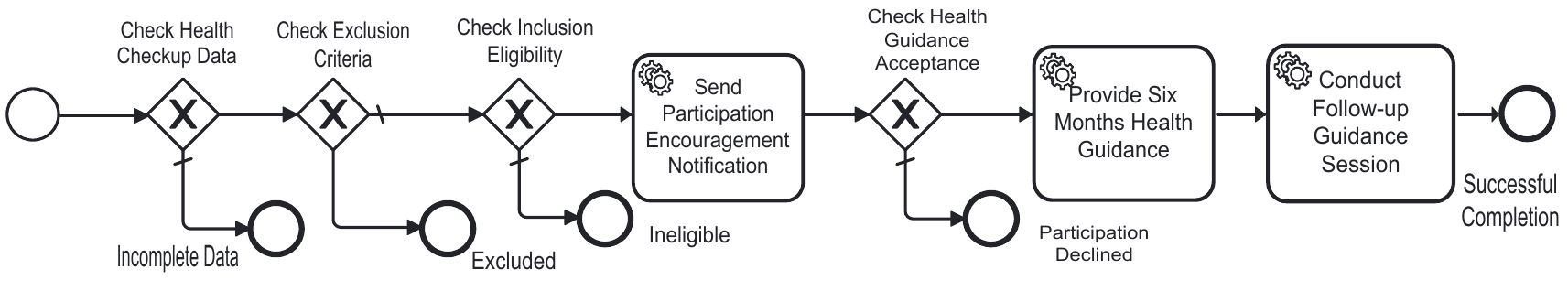}
    \caption{Generated model.}
    \label{fig:city1_bpmn_gen}
  \end{subfigure}\\[8pt]
  \begin{subfigure}[t]{\textwidth}
    \centering
    \includegraphics[width=\textwidth]{%
      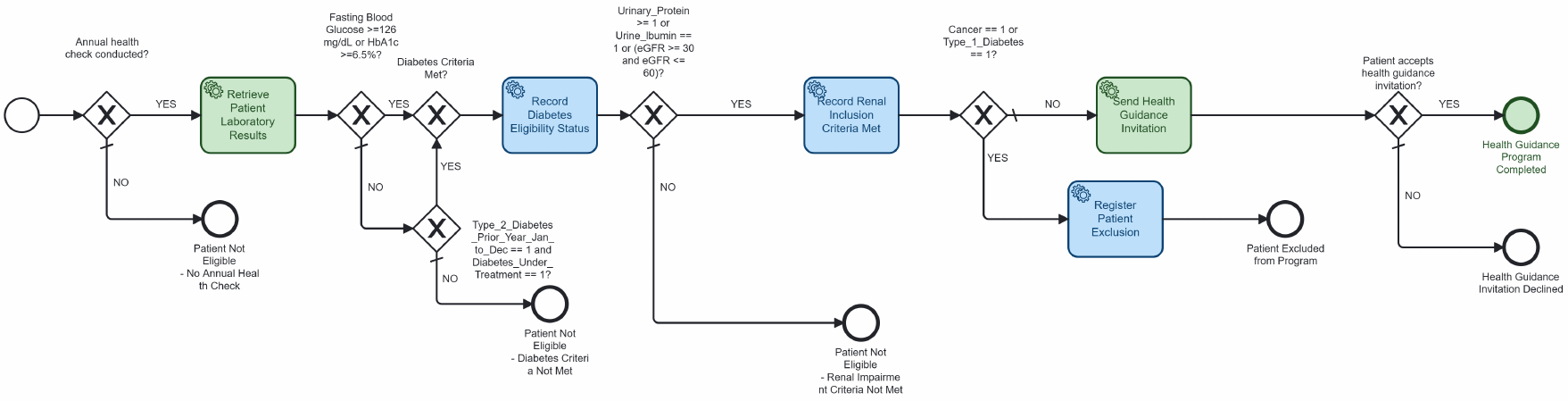}
    \caption{Human-designed baseline.}
    \label{fig:city1_bpmn_human}
  \end{subfigure}
  \caption{BPMN model comparison for {\tt City~1}.  Both models implement
    identical policy logic and produce matching KPI outcomes despite
    differing in gateway topology and task ordering.}
  \label{fig:city1_bpmn}
\end{figure}

Figure~\ref{fig:city1_bpmn} compares a representative generated BPMN
model for {\tt City~1} with the human-designed baseline.  Despite
structural differences, both
implement identical policy logic and produce matching KPI outcomes.

\subsection{City~1: Well-Structured Policy}

\begin{figure}[htbp]
  \centering
  \begin{subfigure}[t]{\textwidth}
    \centering
    \includegraphics[height=6cm]{%
      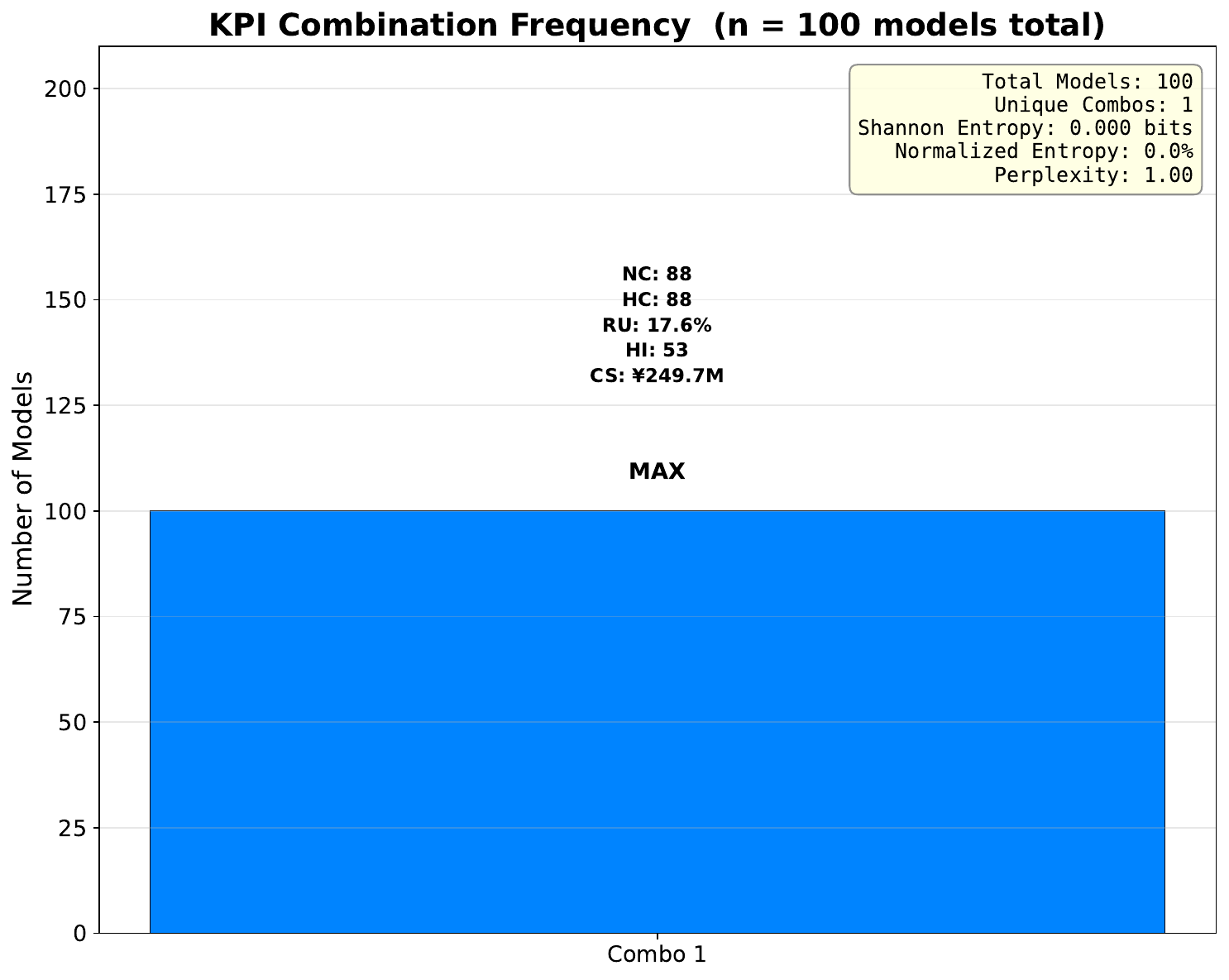}
    \caption{Gemini~2.5 Pro.}
    \label{fig:city1_kpi_pro}
  \end{subfigure}\\[4pt]
  \begin{subfigure}[t]{\textwidth}
    \centering
    \includegraphics[height=6cm]{%
      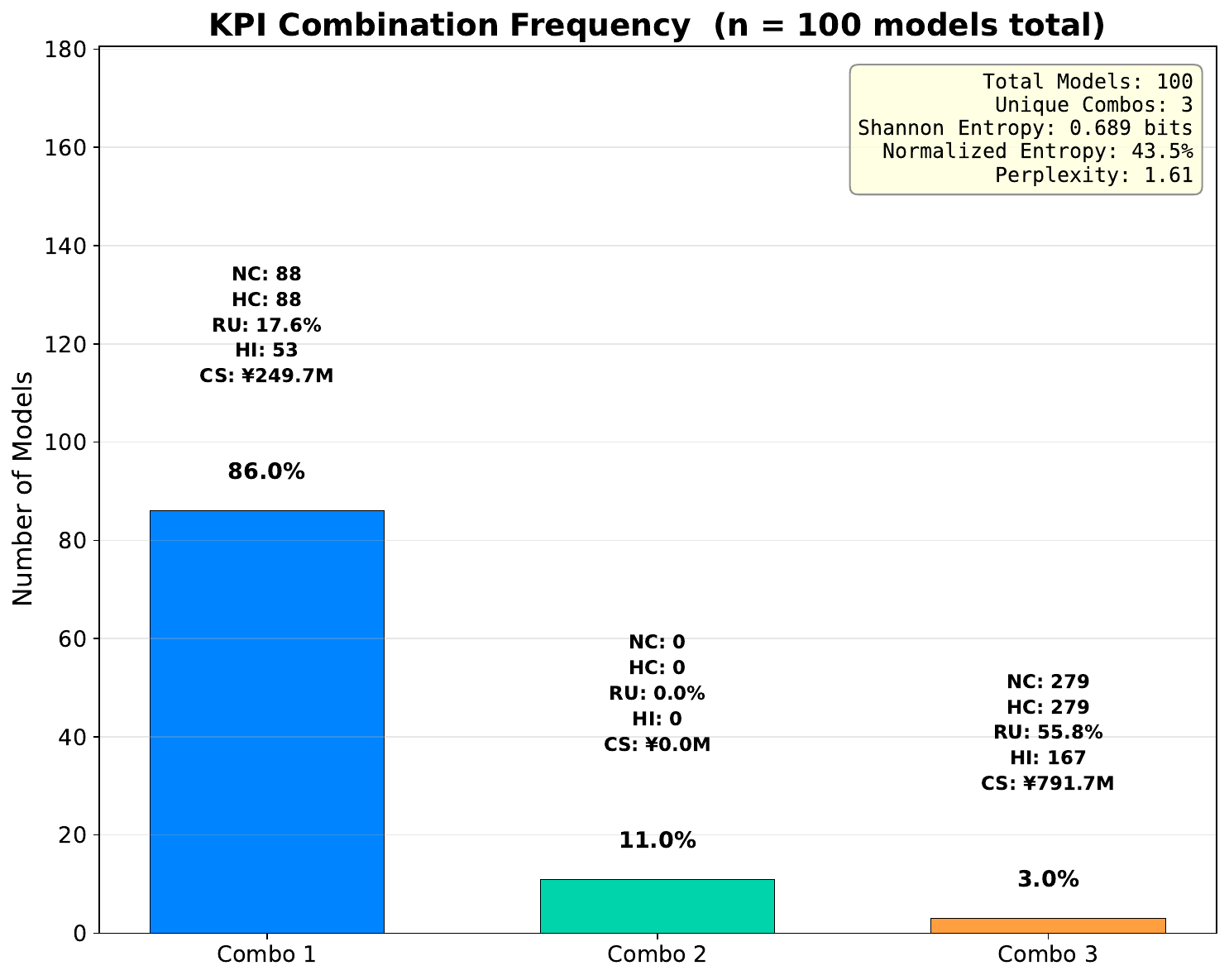}
    \caption{Gemini~2.5 Flash.}
    \label{fig:city1_kpi_flash}
  \end{subfigure}\\[4pt]
  \begin{subfigure}[t]{\textwidth}
    \centering
    \includegraphics[height=6cm]{%
      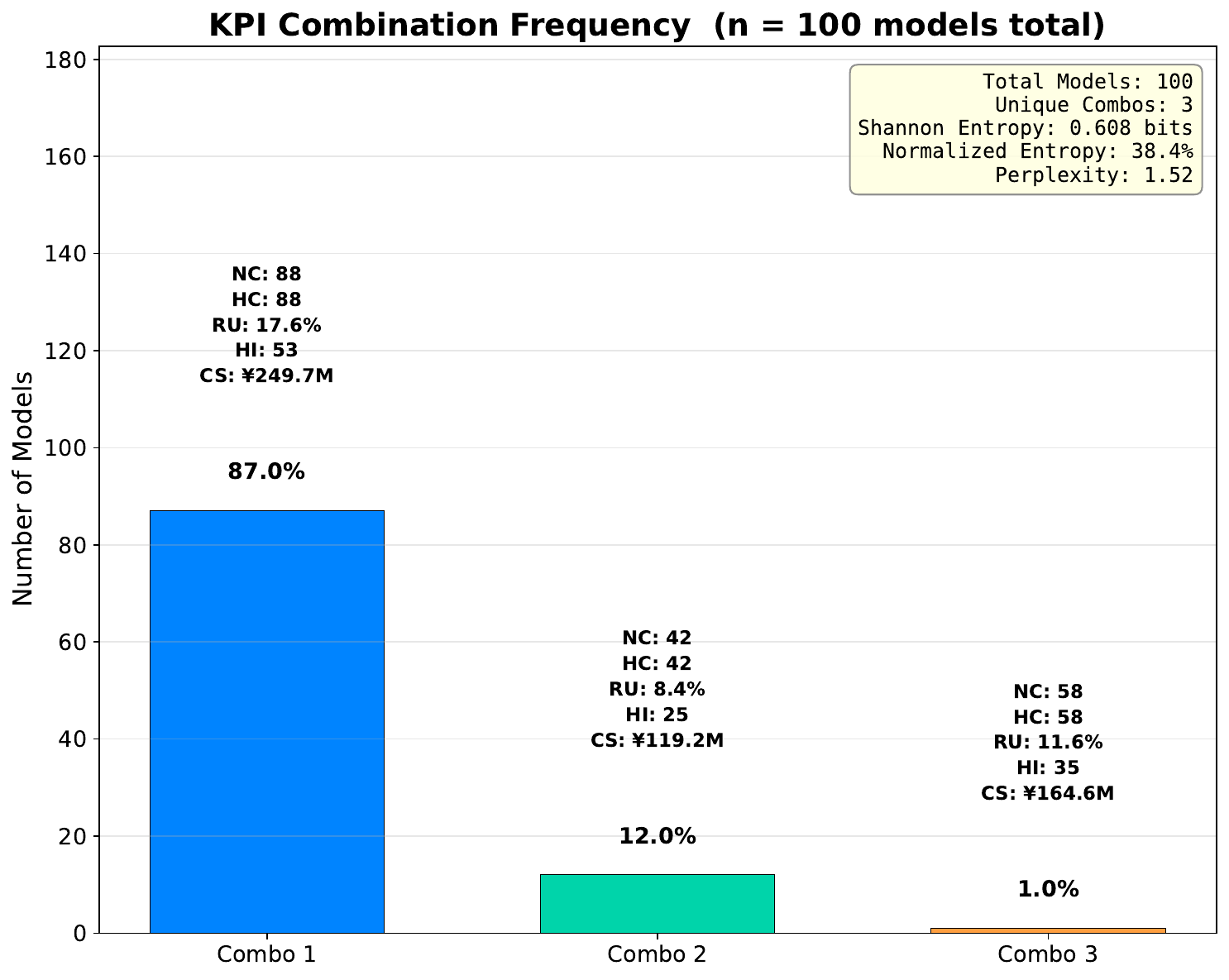}
    \caption{GPT-5.1.}
    \label{fig:city1_kpi_gpt}
  \end{subfigure}
  \caption{Top-5 KPI combination frequency for {\tt City~1} ($n=100$ models
    per LLM).  Pro achieves 100\% ground-truth match with zero entropy.
    Flash and GPT-5.1 each match at 86 - 87\% but exhibit distinct failure
    patterns: Flash produces generation failures (all-zero KPIs); GPT-5.1
    produces variable-binding errors (half-count KPIs).}
  \label{fig:city1_kpi}
\end{figure}

Figure~\ref{fig:city1_kpi} shows the KPI combination frequency for
{\tt City~1} across all three LLMs. Gemini~2.5 Pro
(Figure~\ref{fig:city1_kpi_pro}) achieves complete determinism: all
100~models produce a single KPI combination matching the human baseline
exactly (NC\,=\,88, HC\,=\,88, RU\,=\,17.6\%, HI\,=\,53,
CS\,=\,\textyen249.7M; normalized entropy 0.0\%), confirming that
the pipeline's decision logic is fully determined when the input policy
is unambiguous.

Gemini~2.5 Flash (Figure~\ref{fig:city1_kpi_flash}) and GPT-5.1
(Figure~\ref{fig:city1_kpi_gpt}) achieve ground-truth match rates of
86\% and 87\%, respectively (normalized entropies 43.5\% and 38.4\%),
with three unique combinations each.  The failure modes differ.
Flash produces 11\% generation failures (all-zero KPIs from structural
BPMN violations unrepairable within the iteration budget) and 3\%
over-identifications (NC\,=\,279, from overly broad column selection).
GPT-5.1 produces zero generation failures; its 12\% minority cluster
yields NC\,=\,42 and HC\,=\,42 - roughly half the ground truth - diagnostic
of a variable-binding error in which the LLM selects a narrower database column
instead of the intended disjunction.  The absence of intermediate values
between clusters confirms that this failure mode is binary.

\subsection{City~2: Compound Boolean Eligibility Logic}

\begin{figure}[htbp]
  \centering
  \begin{subfigure}[t]{\textwidth}
    \centering
    \includegraphics[height=6cm]{%
      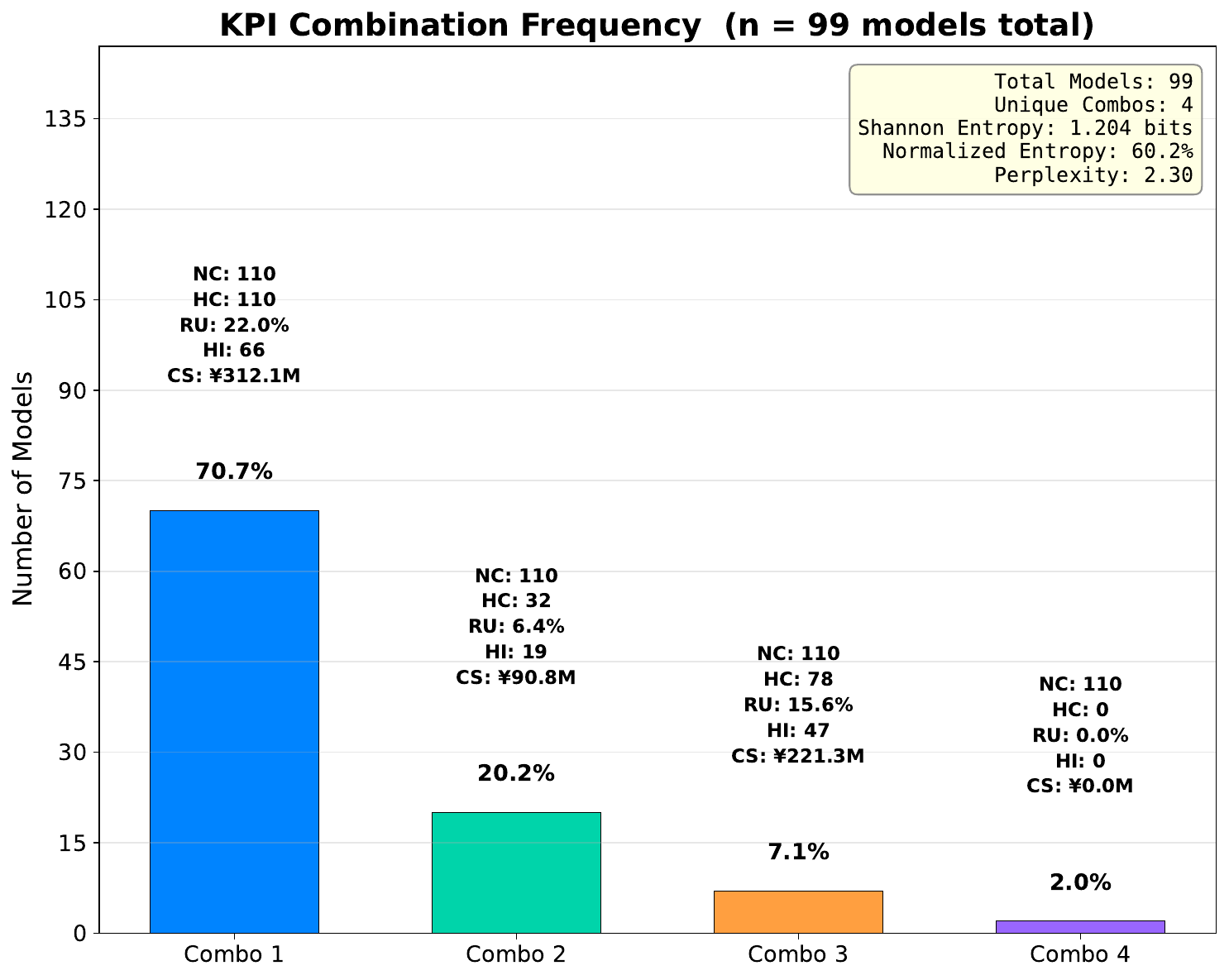}
    \caption{Gemini~2.5 Pro.}
    \label{fig:city2_kpi_pro}
  \end{subfigure}\\[4pt]
  \begin{subfigure}[t]{\textwidth}
    \centering
    \includegraphics[height=6cm]{%
      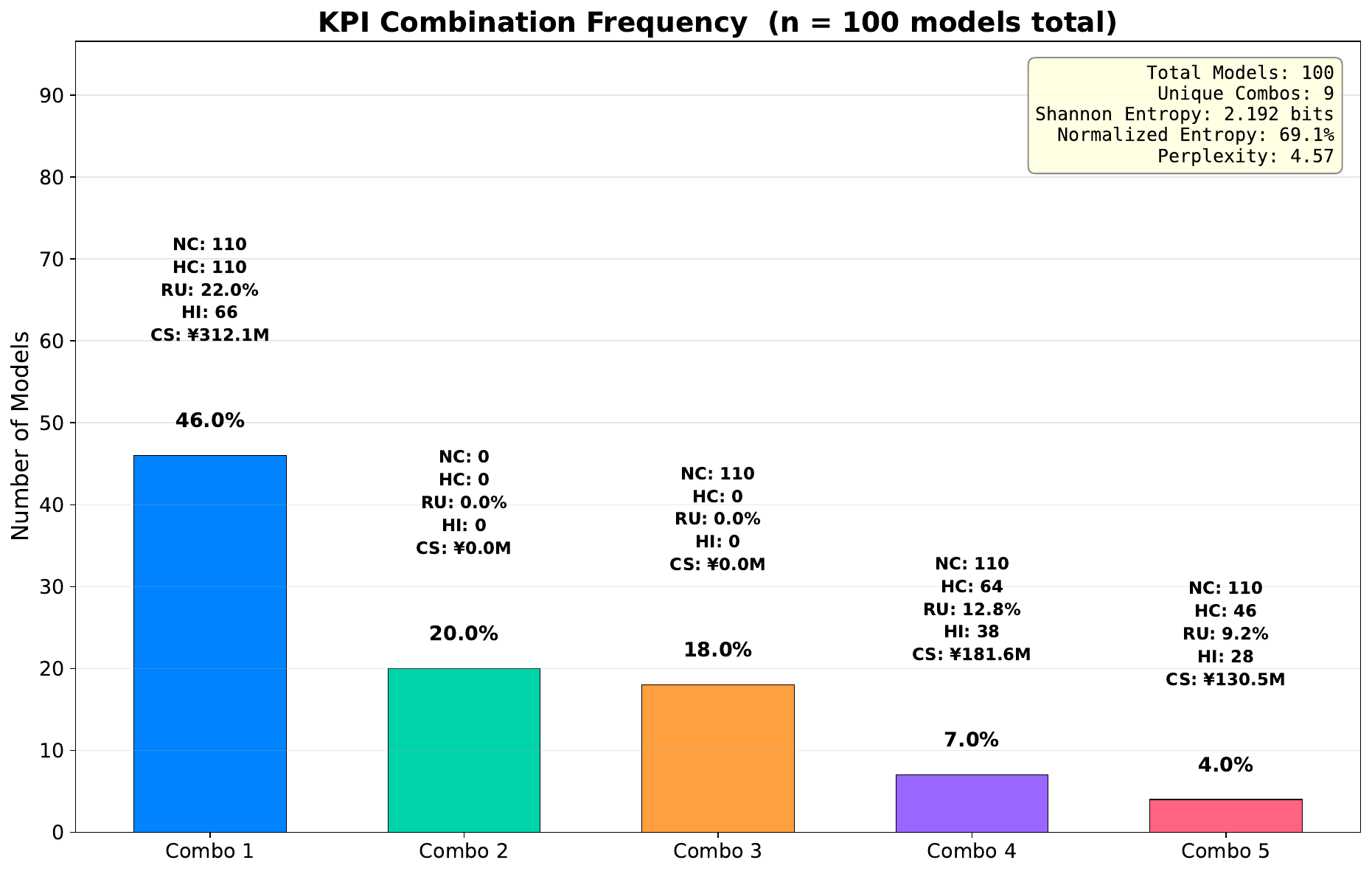}
    \caption{Gemini~2.5 Flash.}
    \label{fig:city2_kpi_flash}
  \end{subfigure}\\[4pt]
  \begin{subfigure}[t]{\textwidth}
    \centering
    \includegraphics[height=6cm]{%
      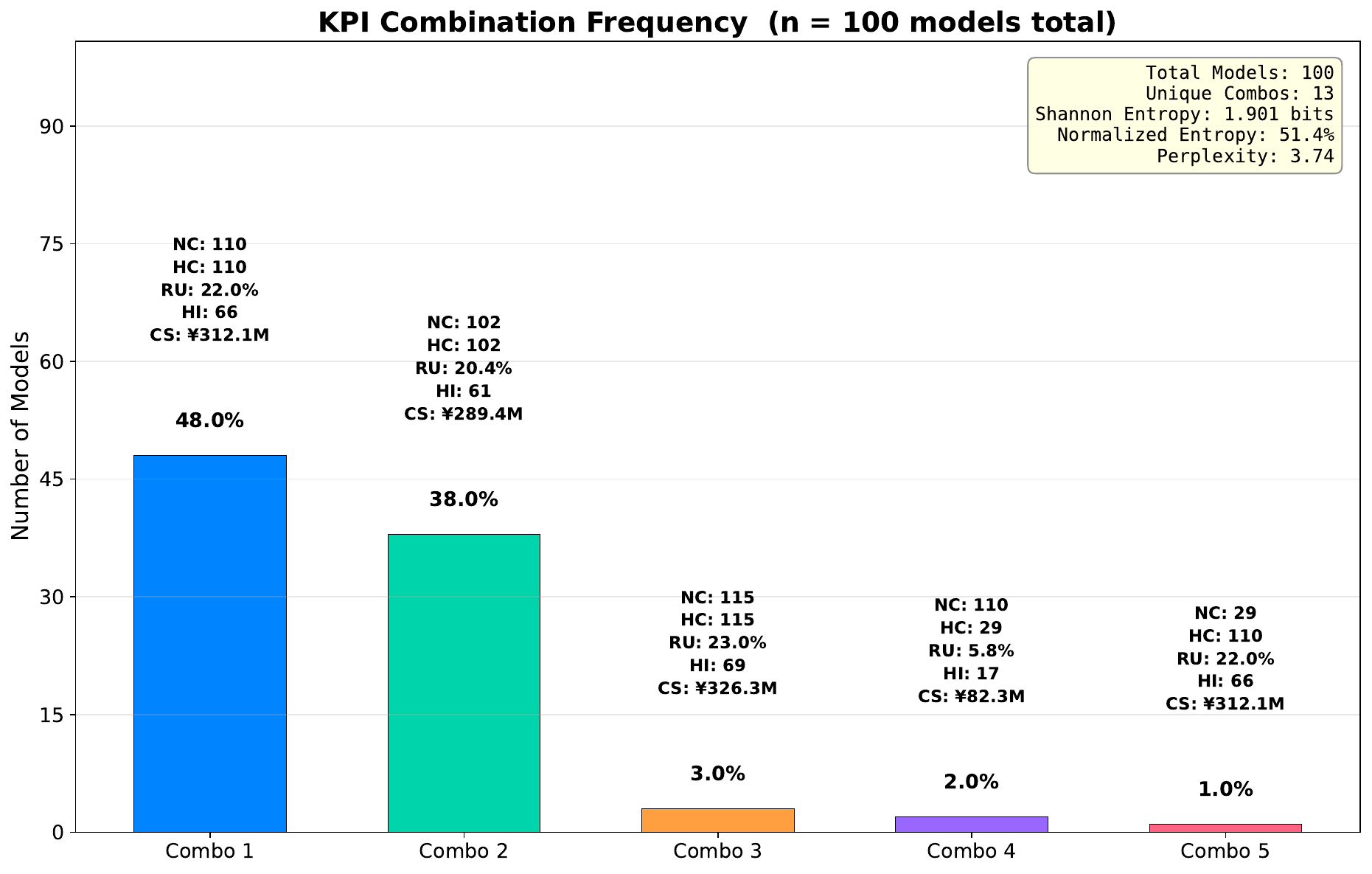}
    \caption{GPT-5.1.}
    \label{fig:city2_kpi_gpt}
  \end{subfigure}
  \caption{Top-5 KPI combination frequency for {\tt City~2}.  Pro maintains
    NC\,=\,110 across all clusters with variation only in HC (normalized
    entropy 60.2\%).  Flash exhibits 20\% generation failures and 69.1\%
    normalized entropy.  GPT-5.1 varies in both NC and HC across 13~unique
    combinations (normalized entropy 51.4\%).}
  \label{fig:city2_kpi}
\end{figure}

{\tt City~2} introduces compound Boolean conditions that combine
disjunctions and conjunctions across four eligibility sub-criteria.
Figure~\ref{fig:city2_kpi} shows the distributions under all three LLMs.

Gemini~2.5 Pro (Figure~\ref{fig:city2_kpi_pro}) produces four distinct
KPI combinations from $n=99$ successfully executing models. The dominant
cluster (70.7\%) matches the ground truth exactly (NC\,=\,110,
HC\,=\,110).  All four clusters share NC\,=\,110 ($\sigma=0$) but
differ in HC (110, 32, 78, and 0; mean $90 \pm 33$), localizing the
disagreement to the guidance-acceptance logic - specifically, the
nested ``(condition~1 OR~2) AND (condition~3 OR~4)''
structure - while the upstream notification pathway is reliably
extracted.  The normalized entropy of 60.2\% reflects structured
disagreement over compound Boolean interpretation.

Gemini~2.5 Flash (Figure~\ref{fig:city2_kpi_flash}) produces nine
unique combinations (normalized entropy 69.1\%, perplexity 4.57) from
$n=100$ models, with a 46\% ground-truth match rate.  Generation
failures account for 20\% of models, and an additional 18\% correctly
identify NC\,=\,110 but assign HC\,=\,0, indicating that Flash
reliably extracts the notification pathway yet frequently fails on
the guidance-acceptance logic.  Minority clusters at HC\,=\,64 (7\%)
and HC\,=\,46 (4\%) further confirm that the compound Boolean
structure drives the variation.

GPT-5.1 (Figure~\ref{fig:city2_kpi_gpt}) produces 13~unique
combinations (normalized entropy 51.4\%, perplexity 3.74).
Its two dominant clusters - 48\% at ground truth and 38\% at a near-miss
(NC\,=\,102, HC\,=\,102) - differ by only eight patients.  Unlike Pro
and Flash, GPT-5.1 also varies in NC ($107 \pm 14$), indicating
occasional misparsing of the notification criterion.

\subsection{City~3: Implicit Temporal Dependencies}

\begin{figure}[htbp]
  \centering
  \begin{subfigure}[t]{\textwidth}
    \centering
    \includegraphics[height=6cm]{%
      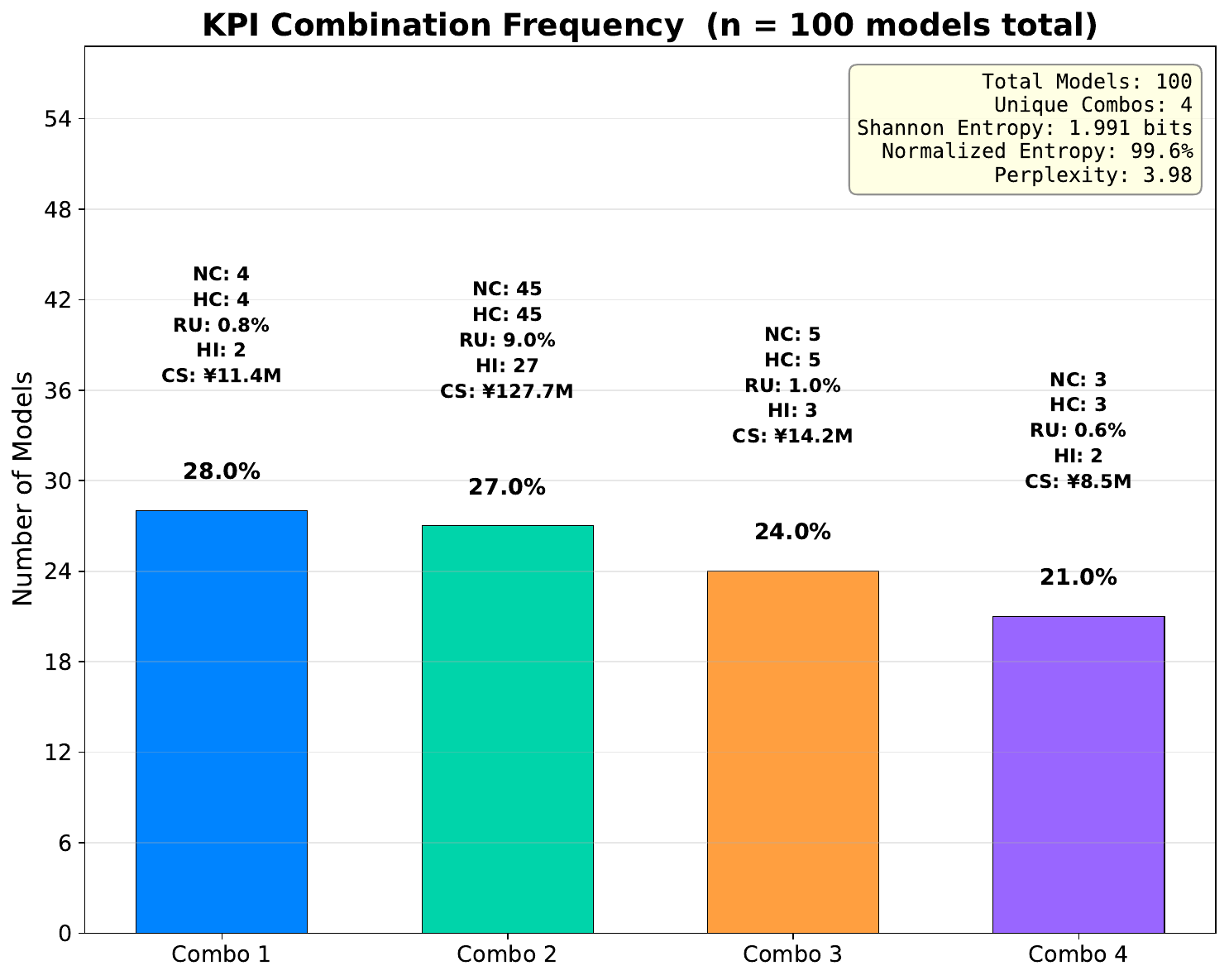}
    \caption{Gemini~2.5 Pro.}
    \label{fig:city3_kpi_pro}
  \end{subfigure}\\[4pt]
  \begin{subfigure}[t]{\textwidth}
    \centering
    \includegraphics[height=6cm]{%
      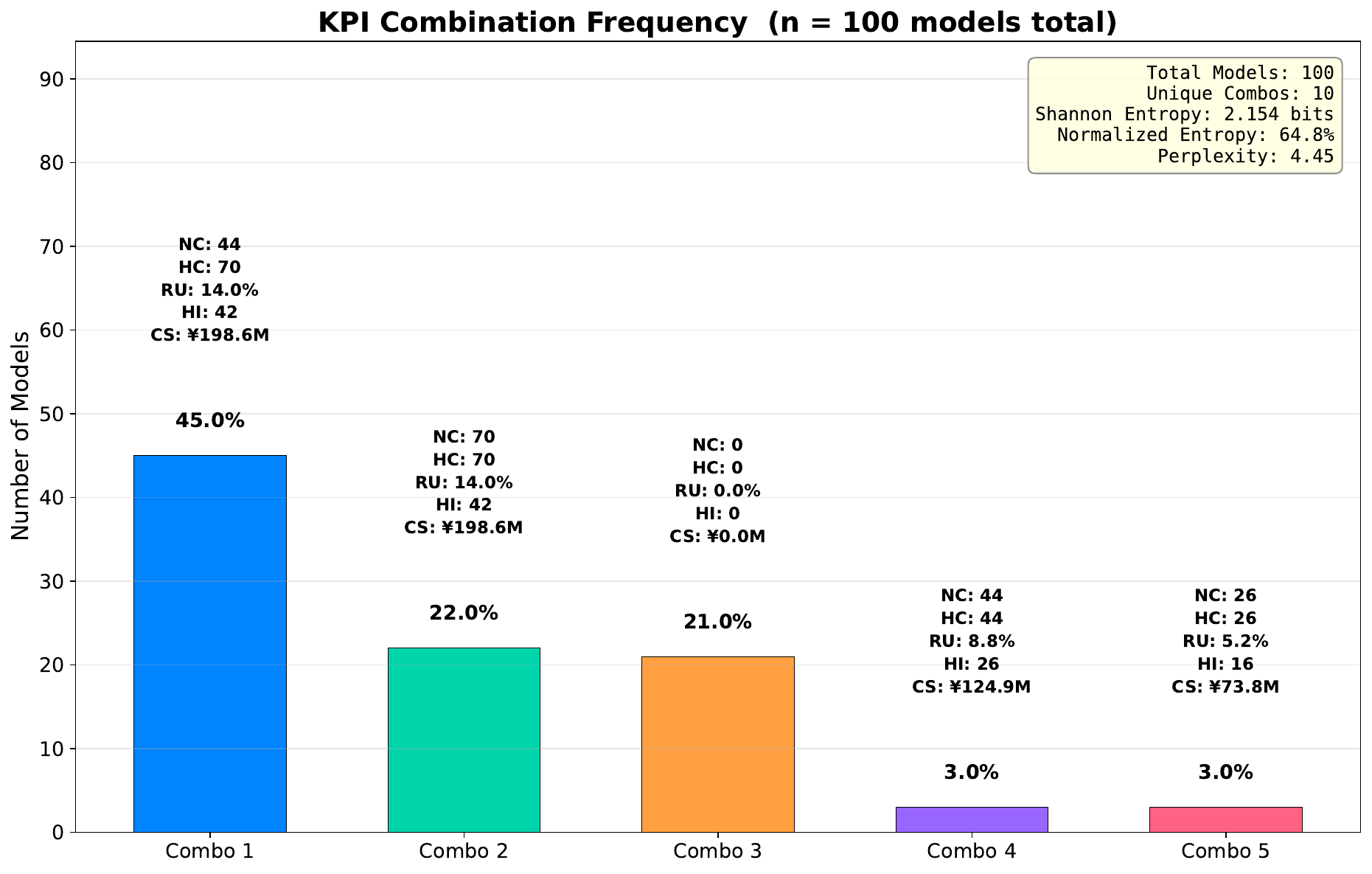}
    \caption{Gemini~2.5 Flash.}
    \label{fig:city3_kpi_flash}
  \end{subfigure}\\[4pt]
  \begin{subfigure}[t]{\textwidth}
    \centering
    \includegraphics[height=6cm]{%
      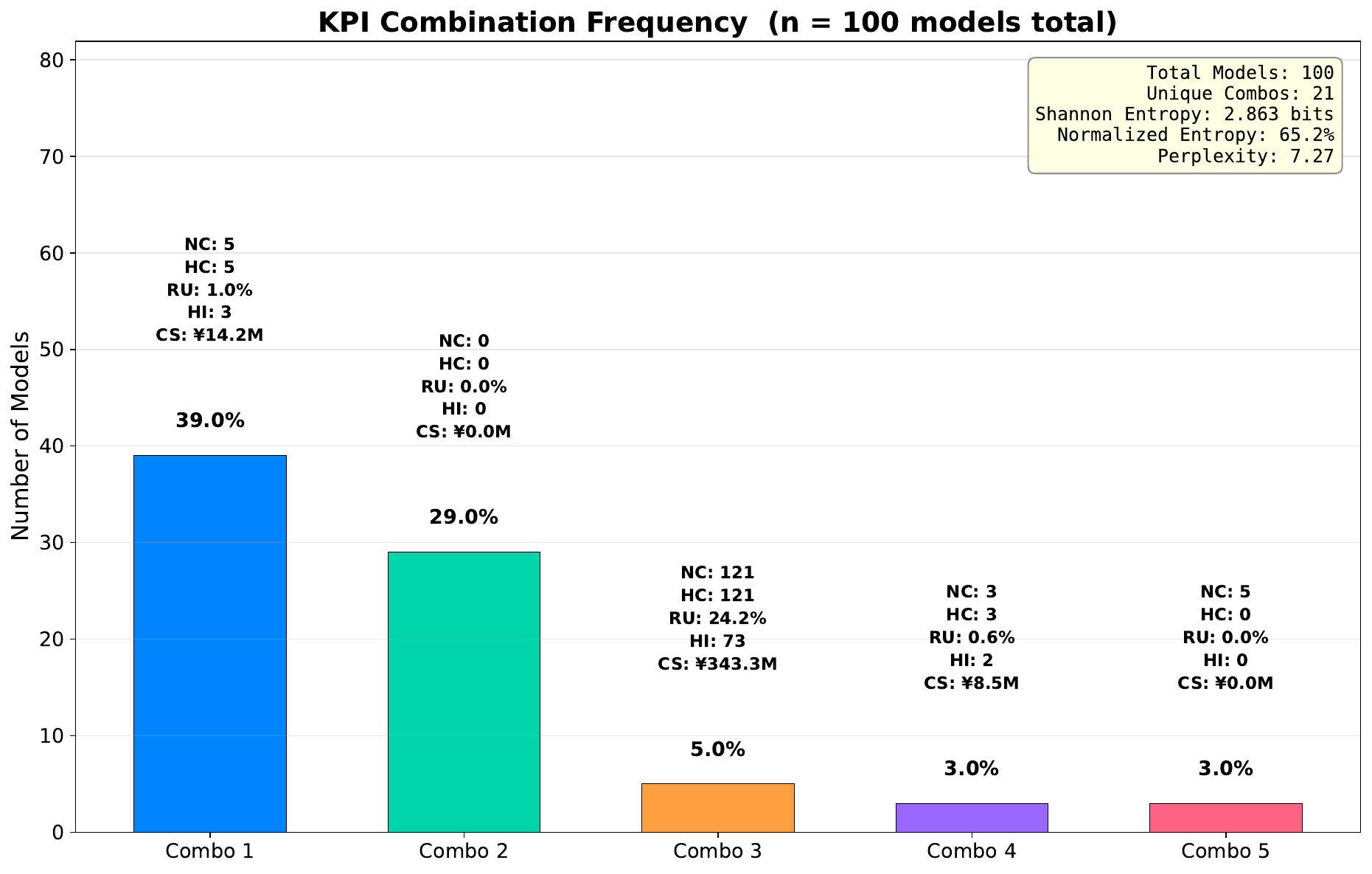}
    \caption{GPT-5.1.}
    \label{fig:city3_kpi_gpt}
  \end{subfigure}
  \caption{Top-5 KPI combination frequency for {\tt City~3} ($n=100$ models
    per LLM).  All three LLMs produce high entropy (Pro: 99.6\%, Flash:
    64.8\%, GPT-5.1: 65.2\%), confirming that the ambiguity is an intrinsic
    document property.  Pro yields four stable clusters; Flash and GPT-5.1
    exhibit long tails with 21-29\% generation failures.}
  \label{fig:city3_kpi}
\end{figure}

{\tt City~3} represents the most challenging case, with eligibility
criteria spanning two fiscal years and implicit sequencing between
screening, consultation, and guidance.  Figure~\ref{fig:city3_kpi}
shows the distributions under all three LLMs.

Gemini~2.5 Pro (Figure~\ref{fig:city3_kpi_pro}) produces four clusters
with near-uniform frequency (28\%, 27\%, 24\%, 21\%), yielding a
normalized entropy of 99.6\% - the theoretical maximum for four outcomes.
NC values across clusters (4, 45, 5, 3) span an order of magnitude,
and the mean-variance analysis confirms multi-modal variation
(NC\,=\,$15 \pm 18$, where $\sigma > \mu$) rather than noise around a
central tendency.

Gemini~2.5 Flash (Figure~\ref{fig:city3_kpi_flash}) produces
10~unique combinations (normalized entropy 64.8\%) with 21\%
generation failures.  GPT-5.1 (Figure~\ref{fig:city3_kpi_gpt}) yields
the highest fragmentation in the evaluation: 21~unique combinations
(normalized entropy 65.2\%, perplexity 7.27) with 29\% generation
failures - the highest rate observed across all experiments.  For
both Flash and GPT-5.1, the standard deviation of NC is either comparable to or exceeds the mean
(Flash: $40 \pm 22$; GPT-5.1: $17 \pm 35$), indicating that the
multi-modal pattern is not specific to Pro.

The critical observation is that \emph{all three LLMs} produce elevated
entropy for {\tt City~3} (99.6\%, 64.8\%, 65.2\%), whereas the same
models produce substantially lower entropy on {\tt City~1} (0.0\%,
43.5\%, 38.4\%).  The inter-city entropy gap far exceeds the inter-LLM
gap within any city, confirming that the elevated uncertainty is an
intrinsic property of the source document rather than an artifact of any
individual model's limitations.

\subsection{Per-Patient Decision Agreement}
\label{sec:per_patient}

The KPI-based evaluation measures aggregate functional equivalence
across only two independent signals (NC and HC), since RU, HI, and CS
are deterministic transforms of HC.  To verify that aggregate agreement
reflects per-patient decision fidelity, we compare each generated
model's eligibility decision against the ground truth on the same 1{,}000
patients and compute agreement rate, F1, recall, balanced accuracy, and
Cohen's $\kappa$.  Table~\ref{tab:agreement} summarizes the results.

For {\tt City~1}, all backends achieve $\geq98.4\%$ agreement and
$\kappa \geq 0.918$, with Pro attaining perfect agreement ($\kappa = 1.000$).
For {\tt City~2}, GPT-5.1 leads ($\kappa=0.931$, 98.9\% agreement); Pro
follows at $\kappa=0.850$; Flash drops to $\kappa=0.615$ due to its
higher generation-failure rate.  For {\tt City~3}, all backends exhibit
near-chance $\kappa$ ($-0.015$ to $0.077$) despite $>92\%$ raw
agreement.  This outcome constitutes an instance of the base-rate paradox: when the
eligible class is rare ($<5\%$), even a model that never selects any
patient achieves high raw agreement while contributing no discriminative
value, as reflected by $\kappa \approx 0$.  The pattern corroborates the
entropy-based findings: high agreement on well-structured policies
degrades when implicit temporal dependencies introduce ambiguity.

\begin{table}[!h]
\centering
\caption{Per-patient decision-agreement metrics (mean across complete variants, 1{,}000 patients each).
Bal.\ Acc.\ $= (\text{Recall} + \text{Specificity})/2$. $^*$Incomplete variants ($<$1{,}000 patients each) are excluded.}
\label{tab:agreement}
\smallskip
\begin{tabular}{llcrrrrr}
\toprule
Model & City & $N$ & Agr.\ (\%) & F1 (\%) & Rec.\ (\%) & Bal.Acc.\ (\%) & $\kappa$ \\
\midrule
  Gemini Flash 2.5 & City 1 & 91$^*$ & 98.4 & 100.0 & 92.3 & 95.6 & 0.918 \\
  Gemini Flash 2.5 & City 2 & 84$^*$ & 94.5 & 93.8 & 60.4 & 79.6 & 0.615 \\
  Gemini Flash 2.5 & City 3 & 84$^*$ & 92.9 & n/a & 0.0 & 46.9 & $-$0.015 \\
  Gemini Pro 2.5 & City 1 & 100 & 100.0 & 100.0 & 100.0 & 100.0 & 1.000 \\
  Gemini Pro 2.5 & City 2 & 99 & 98.0 & 87.4 & 81.6 & 90.8 & 0.850 \\
  Gemini Pro 2.5 & City 3 & 100 & 97.6 & 8.2 & 5.1 & 51.8 & 0.034 \\
  GPT-5.1 & City 1 & 100 & 99.4 & 95.5 & 93.4 & 96.7 & 0.953 \\
  GPT-5.1 & City 2 & 100 & 98.9 & 94.5 & 91.7 & 95.8 & 0.931 \\
  GPT-5.1 & City 3 & 98$^*$ & 97.5 & 13.7 & 16.7 & 57.5 & 0.077 \\
\bottomrule
\end{tabular}
\end{table}

\noindent
Across all three cities, the pattern is consistent: $\kappa$ tracks
the entropy-based uncertainty scores.  {\tt City~1} (low entropy) yields
near-perfect $\kappa$; {\tt City~2} (moderate entropy) yields moderate
$\kappa$ with LLM-dependent variation; {\tt City~3} (high entropy) yields
near-zero $\kappa$ universally.  This concordance between two independent
evaluation axes - aggregate KPI entropy and per-patient decision
agreement - strengthens the conclusion that the pipeline's uncertainty
detector reliably identifies documents whose ambiguity degrades model
fidelity.

\FloatBarrier

\section{Conclusions}
\label{sec:conclusions}

We presented an end-to-end pipeline that transforms healthcare policy documents into executable, KPI-instrumented BPMN
models and evaluated it on diabetic nephropathy guidance policies from
three Japanese municipalities using three LLM backends (Gemini~2.5 Pro,
Gemini~2.5 Flash, and GPT-5.1), generating 100 candidate models per
backend per municipality to characterize output variability.

The pipeline processed all three policies without manual intervention
beyond translation verification (\textbf{RQ1}).  The generated models
ran directly in SpiffWorkflow (\textbf{RQ2}).  LLM-driven KPI
instrumentation automatically linked five KPIs to tasks
(\textbf{RQ3}).  The entropy-based uncertainty detector distinguished
well-specified policies from ambiguous policies: normalized entropy increased
monotonically from {\tt City~1} (0.0\% - 43.5\%) through {\tt City~2}
(51.4\% - 69.1\%) to {\tt City~3} (64.8\% - 99.6\%) across all backends,
with per-patient $\kappa$ dropping from $\geq0.918$ to $\leq0.077$
(\textbf{RQ4}).  Generation failures and variable-binding errors emerged
as distinct failure modes; cross-LLM majority voting on the generated
models can filter the former (non-zero KPI check) and mitigate the
latter by selecting the consensus decision logic.

Limitations include a narrow set of three test cases and a restricted
BPMN construct palette that excludes parallel gateways, timers, and
sub-processes.  Future work will extend the framework to additional
municipalities and non-healthcare domains and will incorporate
human-in-the-loop and automated ambiguity correction for entropy-flagged documents.

\bibliographystyle{splncs04}
\bibliography{references}

\end{document}